\documentclass[conference]{IEEEtran}
\IEEEoverridecommandlockouts
\usepackage{algpseudocode}
\usepackage{algorithm}
\usepackage{amsmath}
\usepackage{amssymb}
\usepackage{balance}
\usepackage{booktabs}
\usepackage[noadjust]{cite}
\usepackage{graphicx}
\usepackage{siunitx}
\usepackage{textcomp}
\usepackage[absolute,showboxes]{textpos}
\graphicspath{{./img/}}
\TPMargin*{3pt}

\newcommand\copyrighttext{
	\footnotesize
	\noindent
	\textcopyright\,2023 IEEE.
	Personal use of this material is permitted.
	Permission from IEEE must be obtained for all other uses, in any current or future media, including reprinting/republishing this material for advertising or promotional purposes, creating new collective works, for resale or redistribution to servers or lists, or reuse of any copyrighted component of this work in other works.}%
\newcommand\copyrightnotice{%
	\begin{textblock*}{7in}(0.75in,0.25in)
		\copyrighttext
	\end{textblock*}
}

\begin{document}

\title{%
Integration of Reinforcement Learning Based Behavior Planning With Sampling Based Motion Planning for Automated Driving%
\thanks{Part of this work was financially supported by the Federal Ministry for Economic Affairs and Climate Action of Germany within the program "Highly and Fully Automated Driving in Demanding Driving Situations" (project LUKAS, grant numbers 19A20004A and 19A20004F).}%
}

\author{%
\IEEEauthorblockN{%
Marvin~Klimke\IEEEauthorrefmark{1}\IEEEauthorrefmark{2},
Benjamin~V\"olz\IEEEauthorrefmark{1}, and
Michael~Buchholz\IEEEauthorrefmark{2}}\\
\IEEEauthorblockA{%
\IEEEauthorrefmark{1}Robert Bosch GmbH, Corporate Research, D-71272 Renningen, Germany.\\E-Mail: {\tt\small\{marvin.klimke, benjamin.voelz\}@de.bosch.com}\\
\IEEEauthorrefmark{2}Institute of Measurement, Control and Microtechnology, Ulm University, D-89081 Ulm, Germany.\\E-Mail: {\tt\small michael.buchholz@uni-ulm.de}%
}}

\maketitle
\copyrightnotice
\bstctlcite{ieeenodash}

\begin{abstract}
Reinforcement learning has received high research interest for developing planning approaches in automated driving.
Most prior works consider the end-to-end planning task that yields direct control commands and rarely deploy their algorithm to real vehicles.
In this work, we propose a method to employ a trained deep reinforcement learning policy for dedicated high-level behavior planning.
By populating an abstract objective interface, established motion planning algorithms can be leveraged, which derive smooth and drivable trajectories.
Given the current environment model, we propose to use a built-in simulator to predict the traffic scene for a given horizon into the future.
The behavior of automated vehicles in mixed traffic is determined by querying the learned policy.
To the best of our knowledge, this work is the first to apply deep reinforcement learning in this manner, and as such lacks a state-of-the-art benchmark.
Thus, we validate the proposed approach by comparing an idealistic single-shot plan with cyclic replanning through the learned policy.
Experiments with a real testing vehicle on proving grounds demonstrate the potential of our approach to shrink the simulation to real world gap of deep reinforcement learning based planning approaches.
Additional simulative analyses reveal that more complex multi-agent maneuvers can be managed by employing the cycling replanning approach.
\end{abstract}

\section{Introduction}
\label{sec:intro}

Automated driving systems are typically composed of several modular components serving well-defined purposes, such as perception, localization, and planning~\cite{saej3131}.
The perception of surrounding road users by a, possibly multi-modal, sensor setup alongside the vehicle's localization yields an environment model (EM) of the traffic scene.
Planning of the future vehicle movement can be divided into high-level behavior planning and subsequent detailed motion planning~\cite{paden2016survey}.
The desired route and the current EM are passed to the behavior planning module, which derives a coarse maneuver specification.
Within motion planning, a drivable trajectory is planned that is finally tracked by local feedback control on the vehicle actuators.

\begin{figure}
	\centering
	\includegraphics[width=\linewidth]{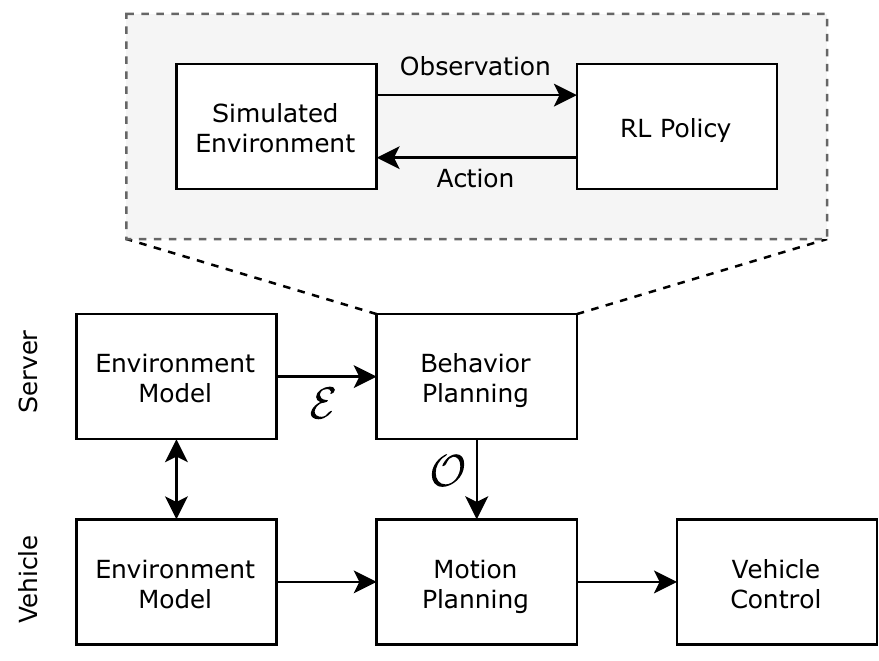}
	\caption{In connected automated driving, the behavior planning module can be moved to a centralized instance, e.g., an edge server. Here, the RL policy interacts with a simulation environment to plan the evolvement of the traffic scene ahead.}
	\label{fig:intro}
\end{figure}

In connected automated driving, this setup allows to move the behavior planning module to a centralized instance, like an edge server, which enables automated intersection management (AIM) \cite{zhong2020autonomous}.
Research of the recent years put forth a large variety of machine learning based approaches to automated driving.
Those methods usually perform integrated behavior planning and motion planning in a single deep neural network, sometimes even incorporating perception or control~\cite{zhu2022survey}.
Training of those networks relies typically on expert demonstrations for performing imitation learning~(IL) or a carefully designed reward function encoding the desired behavior in deep reinforcement learning~(RL).

The present work strives to apply an RL-based policy solely for behavior planning, while keeping motion planning separate.
In this setup, we employ the sampling-based motion planner presented in~\cite{buchholz2021handling} alongside a trajectory tracking controller that could be analytically proven to be safe.
On the behavior planning side, we build upon a learned cooperative planner for urban intersection management in mixed traffic~\cite{klimke2023automatic}.
By leveraging a built-in simulation environment, the evolvement of the traffic scene is rolled out and used for motion specification, as illustrated in Fig.~\ref{fig:intro}.
The core contribution of the current work is fourfold:
\begin{itemize}
\item Proposing a method for the derivation of motion planning objectives from an RL-based planning policy,
\item Exemplary integration of our approach with a sampling-based motion planner,
\item Showing the fundamental feasibility of our approach by deploying it to a real vehicle,
\item Demonstrating its applicability to more complex multi-agent scenarios by simulative analyses.
\end{itemize}

The remainder of the paper is structured as follows:
Section~\ref{sec:sota} gives an overview of the state of the art in machine learning based planning for automated driving.
Afterwards, we introduce our approach for using an RL-based policy in behavior planning and its link to the motion planner in Section~\ref{sec:approach}.
The experimental validation on a real testing vehicle and additional simulative results are presented in Section~\ref{sec:eval}.
Finally, Section~\ref{sec:conclusion} concludes the paper.

\section{Related Work}
\label{sec:sota}

The state of the art in machine learning based methods for automated driving provides a large variety of approaches to planning in urban environments.
While a notable share thereof can be considered end-to-end, i.e., integrating perception, planning, and possibly control into one deep neural network, we focus our literature review on dedicated planning techniques.
For a more extensive overview, the reader is referred to surveys like~\cite{zhu2022survey}.

Training a planning policy for a single ego~vehicle in regular traffic can performed through IL on expert trajectories taken from datasets or simulation~\cite{bansal2019chauffeurnet,chen2019deep,codevilla2018end--end}.
These networks either output direct control signals, like acceleration and steering angle~\cite{codevilla2018end--end}, or a sequence of planned waypoints~\cite{chen2019deep} optionally enriched by a future speed profile~\cite{bansal2019chauffeurnet}.
The latter is then processed by a tracking controller that controls the vehicle's actuators to carefully follow the planned trajectory.
IL approaches share the need for large amounts of ground truth data for training.
While reasonably sized datasets are becoming available, they show regular traffic behavior, but are unsuited for cooperative planning that shall explicitly deviate from priority rules.
Classical RL approaches, like tabular-based ones, are considered out of scope for this work, as deep RL is deemed superior in terms of scalability to complex environments, like robotics~\cite{arulkumaran2017deep}.

Deep RL evades the need for training data by using a simulated environment and a reward signal to rate the actions taken by the policy being trained.
Automated vehicles interact with a dynamically changing number of surrounding road users, which requires a suited input representation for effective learning.
Common approaches include rendering a top-down raster image of the semantic scene~\cite{capasso2021end--end,maramotti2022tackling}, using an aggregation function over variable-sized inputs~\cite{huegle2019dynamic}, and graph-based representations~\cite{huegle2019dynamic,hart2020graph}.
Our prior works propose to use a graph-based representation and a suited graph neural network (GNN) for cooperative multi-agent planning at urban intersections~\cite{klimke2022cooperative,klimke2022enhanced}.
Recently, this model has been extended to conduct cooperative planning in mixed traffic, i.e., the simultaneous road usage by human-driven vehicles and automated vehicles~\cite{klimke2023automatic}.
All those RL-based approaches have in common that they require immediate state feedback on a selected action.
The current observation is followed by an immediate action, which is only valid for this time step and then superseded by the next chosen action.
This action space is thus not suited for combination with a tracking controller that requires a trajectory planned ahead.

Those limitations might be one reason why there are only very few RL-based automated driving systems deployed to real world.
In~\cite{kendall2019learning}, an RL approach was used for lane following using the on-board computation system of the real testing vehicle.
Moreover, an RL policy was trained for roundabout handling in simulation and subsequently deployed to a real testing vehicle under controlled conditions~\cite{maramotti2022tackling}.
Both works do not consider interaction with other road users in the real-world experiments.
The current state of the art in RL-based planning reveals a significant \emph{simulation to real world gap}.

\section{Proposed Approach}
\label{sec:approach}

In this section, we present our approach to employ an RL policy for behavior planning in automated driving.
Subsection~\ref{ssec:problem} starts by concisely defining the problem at hand.
Afterwards, the interface specifications are discussed in Sec.~\ref{ssec:interfaces} and the core algorithm for deriving motion planning objectives is introduced in Sec.~\ref{ssec:derivationmpobjective}.
Finally, Sec.~\ref{ssec:learningmodel} briefly introduces the RL model from our previous works that is used throughout the study and discusses minor changes that became necessary to derive viable tasks for the motion planner.

\subsection{Problem Statement}
\label{ssec:problem}

The behavior planning module in cooperative automated driving is tasked with deriving a high-level maneuver plan for all involved vehicles.
A complete server-side EM~$\mathcal{E}$ serves as the basis for planning the cooperative maneuver (cf. Fig.~\ref{fig:intro}).
Each connected automated vehicle (CAV) that takes part in the cooperation runs its own motion planning (MP) instance, which has to be supplied with a viable MP objective $\mathcal{O}$, i.e., a high-level plan.
To enable the detailed planning of a smooth and drivable trajectory, the MP objective must specify the desired motion for a sufficient planning horizon (at least multiple seconds) into the future.
The RL policy, on the other hand, only provides an instantaneous action for the next time step.
After the state update, a new observation is passed to the policy, which then selects the next action and the process repeats.

The iterative nature of the RL policy, requiring immediate state feedback, and the demand for advance planning of the MP algorithm pose an inherent gap.
The present work strives to close this gap by using a built-in simulation environment to roll out the initial traffic scene.

\subsection{Interfaces}
\label{ssec:interfaces}

The server-side EM might be fused from multiple data sources, e.g., cooperative awareness messages (CAM, \cite{etsi2014cam}), collective perception messages (CPM, \cite{etsi2019cpm}), and infrastructure sensors~\cite{buchholz2021handling}.
Due to the variety of data sources, the server-side EM is assumed to contain all vehicles that might be relevant for planning.
In the current setup, vulnerable road users, like pedestrians or bicyclists, are not considered.
In the EM, vehicle $\nu_{\mathrm{id}}$ is denoted by the tuple
\begin{equation}
(\mathrm{id},\,\boldsymbol{T}_{\mathrm{id}},\,v,\,\mathcal{D},\,c) \in \mathcal{E},
\label{eq:em}
\end{equation}
where $\mathrm{id}$ denotes a unique integer identifier, and the binary flag $c$ indicates whether the vehicle is controllable, i.e., is willing to cooperate during the maneuver.
The vehicle pose is described by $\boldsymbol{T}_{\mathrm{id}}$, while $v$ denotes its speed.
Moreover, $\mathcal{D}$ contains the desired route or destination of the vehicle, which is provided via vehicle-to-infrastructure communication but might be unknown for non-connected regular vehicles.
As long as not noted otherwise, all poses are denoted in a local east-north-up (ENU) frame.

The MP interface is based on the \emph{SuggestedManeuverContainer} of the maneuver coordination protocol proposal in \cite{mertens2021extended}.
For the present work, we extend the fields of the maneuver coordination message (MCM) by additional information, which would be inferred from a shared map in practice.
Thus, the MP objective is defined as
\begin{equation}
\mathcal{O} = (\mathcal{P},\,\boldsymbol{v}^{+},\,\mathrm{AP}),
\label{eq:mpobjective}
\end{equation}
where $\mathcal{P}$ describes the path specification, e.g., as a series of centerline waypoints and $\boldsymbol{v}^{+}$ the maximum speed profile along this path.
More precise longitudinal control is imposed by anchor points, comprised of tuples
\begin{equation}
\mathrm{AP} = (\boldsymbol{p}_\mathrm{AP},\,t_\mathrm{AP},\,v_\mathrm{AP}),
\end{equation}
where $\boldsymbol{p}_\mathrm{AP}$ denotes the position of this anchor point.
It shall be crossed by the vehicle at time $t_\mathrm{AP}$ with a desired speed of~$v_\mathrm{AP}$.

\subsection{Derivation of Motion Planning Objectives}
\label{ssec:derivationmpobjective}

Given the current EM state and an RL policy, MP objectives are derived using a built-in simulator.
Therein, the RL policy is employed to predict the scene evolvement based on the environment model as the initial state.
Subsequently, the simulated vehicle trajectories are used to derive MP objectives for all CAVs.

Algorithm~\ref{alg:derivempobjective} depicts the core idea of our approach.
We employ the RL training and evaluation environment from~\cite{klimke2022enhanced} that is based on the open-source simulator Highway-env~\cite{highway-env}.
The RL policy maps an observation $o \in \Omega$ provided by the simulation environment and selects a joint action $a$ from the longitudinal acceleration space~$A$.

\begin{algorithm}
\caption{Derive MP objectives from initial EM state.}
\label{alg:derivempobjective}
\begin{algorithmic}[1]
\Require Server-side EM $\mathcal{E}$, RL planning policy $\pi: \Omega \rightarrow A$.
\Ensure MP objective for each vehicle $\mathcal{O}_\nu$.
\State $\mathrm{sim} \leftarrow \mathrm{initialize~simulator}$
\ForAll {$\nu \in \mathcal{E}$}
	\State $\mathrm{sim.addVehicle}(\boldsymbol{T}_\nu,\, v_\nu,\, \mathcal{D}_\nu)$
\EndFor
\Repeat
	\State $o \leftarrow \mathrm{sim.observe()}$
	\State $a \leftarrow \pi(o)$ \# select next action
	\State $\mathrm{sim.step}(a)$
\Until {all vehicles reached destination \textbf{or} timeout}
\ForAll {$\nu \in \mathcal{E}~\mathbf{if}~c_\nu~\mathrm{is~true}$}
	\State $\mathcal{O}_\nu \leftarrow \mathrm{sim.getObjective}(\nu)$
\EndFor
\State \Return all $\mathcal{O}_\nu$
\end{algorithmic}
\end{algorithm}

A planning run is triggered whenever a new vehicle enters the operational area (e.g., on an intersection access) that was previously not considered in the maneuver.
In this case, the plan has to be adapted immediately to consider any further conflicts emerging due to the additional vehicle.
Moreover, cyclic replanning can be issued to quickly react to vehicles not obeying to the initial plan.
With presence of human-driven non-connected vehicles, this approach might become imperative due to the ambiguities in human drivers' intentions.
Once a planning run is requested, all vehicles in the environment model are instantiated in the simulator according to their current state.

Subsequently, the built-in simulator predicts the traffic scene at \SI{5}{\hertz}, which seems to be a good trade-off between responsiveness and runtime considerations.
In each cycle, the current state is observed and passed to the trained RL policy, which selects a longitudinal acceleration for each vehicle.
The Highway-env lateral controller sets the steering angle to guide the vehicle on its desired route provided by all CAVs.
For regular vehicles that do not take part in the cooperative maneuver, a worst-case route estimate is used.
Thereby, all potential conflict points are considered, as long as the true maneuver decision is unknown.
Once such a potential conflict point can be ruled out, replanning is triggered to optimize the cooperative plan.
A kinematic bicycle model~\cite{kong2015kinematic} is used to move the vehicles to their next position.

During the simulation, the simulated trajectories of all vehicles are buffered for later use.
The simulation ends either when a timeout triggers or as soon as all vehicles have reached their destination, e.g., they passed the intersection area.
Aborting by timeout is beneficial to limit the response time in very crowded scenarios.
However, it requires cyclic replanning as the MP objective does not necessarily guide the vehicle the whole way to its destination.
In practice, it rarely takes more than one second to derive a full maneuver from the RL policy, i.e., until all vehicles reached their destination.

\begin{figure}
	\centering
	\includegraphics[width=\linewidth]{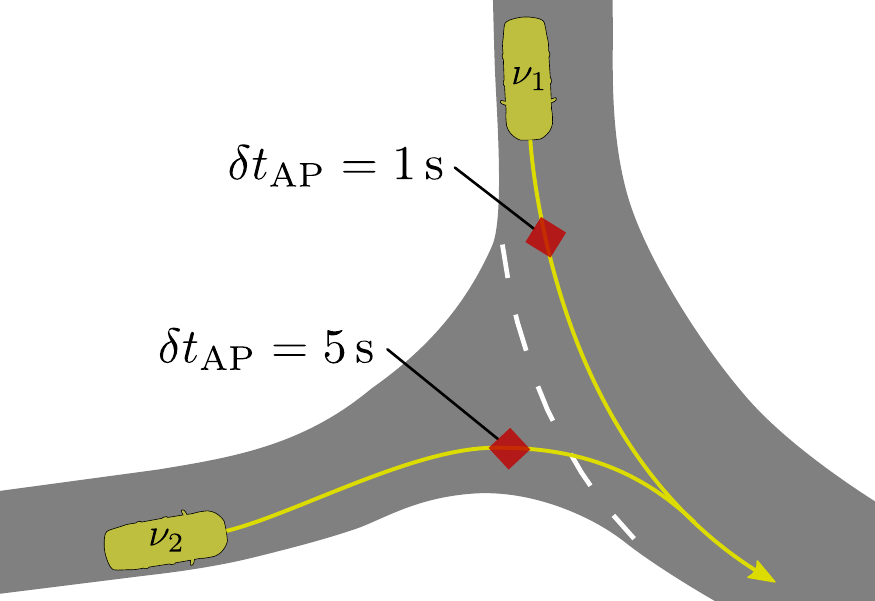}
	\caption{Two automated vehicles on conflicting paths (yellow). By setting one anchor point (red rhombus) for each vehicle, the intersection traversal can be coordinated by setting appropriate anchor times $\delta t_\mathrm{AP}$.}
	\label{fig:ap_placement}
\end{figure}

Once the simulation terminates, the MP objective for each vehicle is derived from the recorded simulated trajectory.
Note that no velocity reduction due to curvature is necessary, as the motion planner handles this on its own.
The proposed approach allows for nearly arbitrary anchor point placement.
However, we argue that the anchor point should be placed close to the conflict points, to minimize the effect of prediction uncertainties.
In the present work, one anchor point per vehicle is placed at the intersection entry on the respective access lane, as illustrated in Fig.~\ref{fig:ap_placement}.
No additional anchor points are used to give the motion planner as much room for trajectory optimization as possible.
In other words, we do not care about how the vehicle approaches the intersection, as long as it meets its space-time slot at the intersection.
The anchor speed $v_\mathrm{AP}$ is set to the simulated vehicle's speed at the anchor position.
Due to implementation details, the timing information is provided as the duration~$\delta t_\mathrm{AP}$ between the current MP objective timestamp and the designated anchor time~$t_\mathrm{AP}$.
Thereby, also the latency between EM perception and output of the MP objective is compensated for.

\subsection{Learning Model}
\label{ssec:learningmodel}

In this paper, we use our RL model for cooperative behavior planning from~\cite{klimke2023automatic}, which is mainly an extension of \cite{klimke2022enhanced} to mixed traffic.
This section briefly introduces the key properties of the learning model that are most relevant to the interaction with motion planning algorithms.

\begin{figure}
	\centering
	\includegraphics[width=\linewidth]{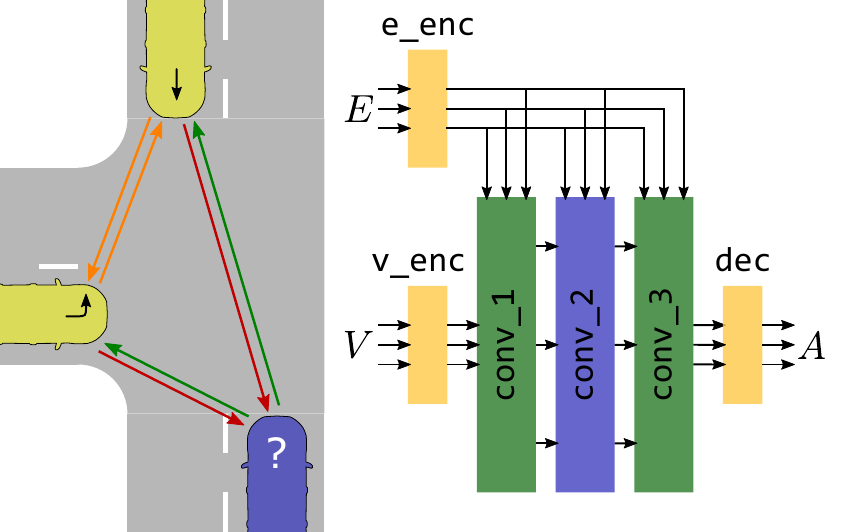}
	\caption{Overview on the learning model for cooperative planning in mixed traffic from~\cite{klimke2023automatic}. Each vehicle in the traffic scene is encoded as a graph vertex in $V$. Conflict relations between vehicles are denoted as edges in $E$. A graph neural network, composed of different message passing layers is trained to select an action from $A$.}
	\label{fig:model}
\end{figure}

The core of the model is a graph-based scene representation that maps each vehicle to a graph vertex in the set $V$, as depicted in the left of Fig.~\ref{fig:model}.
The vertices carry a small set of input features, like position and speed of the corresponding vehicle.
Nodes representing vehicles that drive on conflicting paths are connected by edges that denote the type of conflict relation in the set $E$.
The representation also supports the encoding of pairwise features, like distances and relative priorities.

In the right half of Fig.~\ref{fig:model}, the neural network architecture of the \emph{actor} network is depicted.
The policy is trained using the TD3~\cite{fujimoto2018addressing} actor-critic RL algorithm.
The accompanying \emph{critic} network architecture only deviates slightly from the actor on the output side~\cite{klimke2022cooperative} and is omitted for brevity.
Both, vertex features and edge features are first encoded using fully connected layers (yellow).
Afterwards, the features are passed through three GNN layers that subsequently update the vertex features.
The output decoder infers one joint action in $A$, i.e., one commanded acceleration per vehicle, from the resulting vertex features.
Note that there is no inherent restriction on the dimensionality of the representation and may change arbitrarily.
Vehicles meanwhile joining or leaving the scene can be considered effortlessly, which is a key benefit over alternative representations, like fixed-size vectors or raster images.

The RL algorithm is trained in a simulation environment based in the open-source Highway-env~\cite{highway-env}, which was enhanced for multi-agent planning~\cite{klimke2022cooperative}.
The reward function proposed in our previous works encourages the cooperative planning policy to optimize for the average speed over all vehicles, including those that are not under command.
Simulated manually driven vehicles behave according to regular priority rules and thereby train the policy for planning in mixed traffic.
Although the policy itself is not exposed to measurement uncertainties during training, it proved to be robust against typical uncertainties induced by imperfect measurement as shown in~\cite{klimke2023automatic}.

During the development of the MP link approach presented in the present work, some minor improvements to the learning model became necessary that shall be described in the following.
First, the acceleration regularization through a reward component was dropped in favor of a smaller acceleration range.
It became apparent, that the maximum acceleration of $\pm\SI[per-mode=reciprocal]{5}{\meter\per\second\squared}$ yields a too agile behavior, despite the penalization of large acceleration magnitudes in the reward function.
In our experiments, an acceleration range of $\pm\SI[per-mode=reciprocal]{3}{\meter\per\second\squared}$ leads to drivable maneuvers.
Still, the full acceleration magnitude is seldomly used because smooth driving generally achieves higher rewards.
Additionally, the reward function was modified to encourage the RL policy to keep larger safety margins on the intersection.
The legacy proximity penalty triggered only for very close encounters.
In this work, we doubled the penalizable distance and reduced the gradient accordingly to keep the reward component weighting equal.

Finally, the handling of velocity limits in curved lanes is tweaked.
The legacy maps only contained a strict speed limit for each lane segment, compliance with which is ensured by an intermediate controller.
In case of heavily curved lanes (e.g., for $\nu_2$ in Fig.~\ref{fig:ap_placement}), a significant jump in the speed limit would be required at the lane segment transitions.
We propose to additionally introduce a \emph{speed advice} quantity, which anticipates the curvature ahead of the vehicle and provides a smooth velocity profile to be tracked by the controller.
This feature yields a more reasonable driving behavior in the built-in simulator, without requiring the model to explicitly learn the correct curve speed.
Thus, we retain the independence of concrete intersection geometry within the learning model, which was one of the core architectural choices in~\cite{klimke2023automatic} and prior works.

\section{Experiments}
\label{sec:eval}

To the best of our knowledge, this work is the first to apply an RL policy dedicatedly for behavior planning in automated driving while employing a separate motion planner.
Thus, there is no benchmark or baseline approach available to compare against.
Instead, we opt for deploying our approach to a real testing vehicle and demonstrate its feasibility through driving maneuvers on a test track.
Section~\ref{ssec:setup} introduces the maneuver setup for our real vehicle analysis.
Afterwards, different modes of operation with respect to scheduled replanning are investigated in Sec.~\ref{ssec:replanning} using the real vehicle recordings.
Finally, the evaluation is extended by multi-agent planning results that were obtained in simulation.

\subsection{Experiment Setup}
\label{ssec:setup}

\begin{figure}
	\centering
	\includegraphics[width=\linewidth]{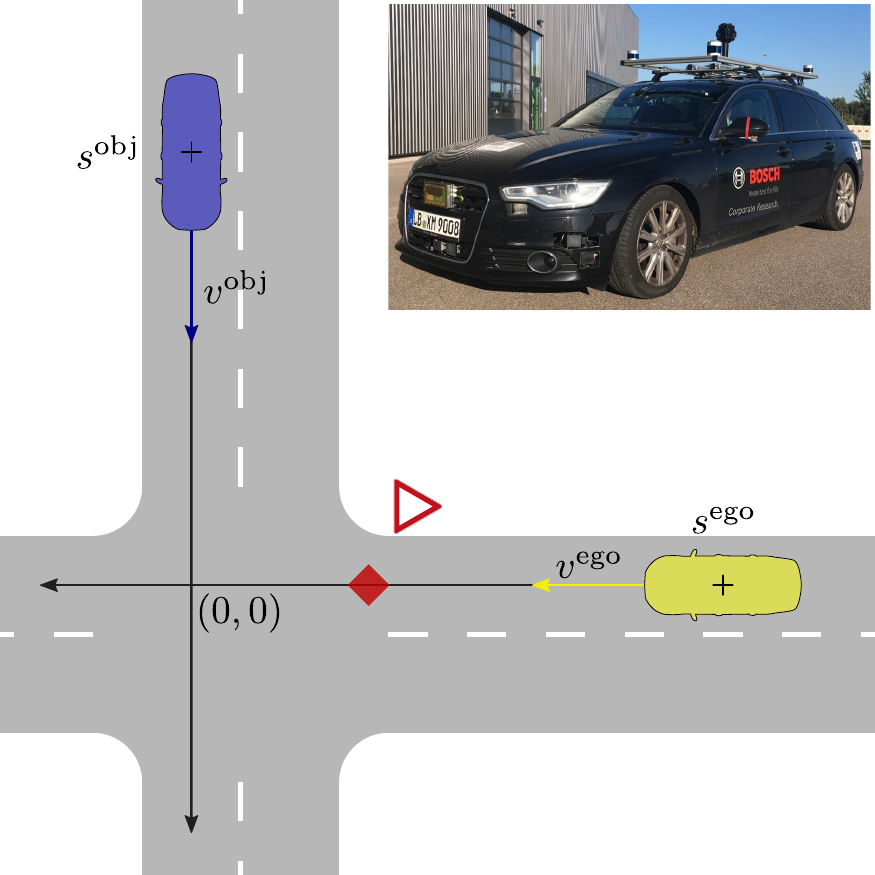}
	\caption{Maneuver setup for the test with a real automated vehicle (yellow) and a simulated prioritized regular vehicle (blue). The vehicles' positions relative to the conflict point are denoted $s^\mathrm{ego}$ and $s^\mathrm{obj}$ for the automated vehicle and object vehicle, respectively. One anchor point is placed at the intersection entry (red rhombus). The testing vehicle used throughout this study is depicted in the upper right.}
	\label{fig:maneuver}
\end{figure}

Figure~\ref{fig:maneuver} depicts the experiment setup comprised of one automated (AD) vehicle and one crossing regular vehicle, which is prioritized over the AD vehicle.
The maneuver is conducted in a vehicle-in-the-loop setting, with a real testing vehicle constituting the ego vehicle, while the crossing regular vehicle is simulated concurrently.
For the current study, the RL~planner was executed on the testing vehicle for simplicity.
This setup allows us to demonstrate edge cases that would be unsafe to conduct with two real vehicles.
We restrict the maneuver to straight driving and vary the initial velocities $v^\mathrm{ego}$, $v^\mathrm{obj}$ and positions $s^\mathrm{ego}$, $s^\mathrm{obj}$ of both vehicles.
For the AD vehicle, the initial position refers to the point, where the RL planner takes over control.

Our proposed approach was deployed to an Audi~A6 testing vehicle (depicted in the upper right of Fig.~\ref{fig:maneuver}), which accepts longitudinal (acceleration) and lateral (steering) commands.
The derived MP objectives are processed by the sampling-based motion planner presented in~\cite{buchholz2021handling}, which plans a drivable trajectory.
Note that all collision and safety checks within the motion planner are turned off so that potential shortcomings of the maneuver become apparent in the result.
A low-level tracking controller finally provides the actuating variables guiding the vehicle on the planned trajectory.

\subsection{Analysis on Replanning}
\label{ssec:replanning}

In Sec.~\ref{ssec:derivationmpobjective}, multiple options for triggering a planning run have been introduced.
In the simplest case, a maneuver is planned only when an additional, previously not considered vehicle, appears.
The number of vehicles in the crossing maneuver from Fig.~\ref{fig:maneuver} is constant and thus only a single planning run is triggered.
Thus, we refer to this mode as \emph{single-shot planning} throughout this section.
Should a vehicle not obey to the planned maneuver, which might be the case especially when regular vehicles are present, it may become infeasible during execution.
Such states must be avoided because dangerous situations may arise.
The RL planner integration supports \emph{cyclic replanning} to quickly react to changes in maneuver evolution.
To ensure that the vehicles' reactions to the most recent MP objective are visible in the next planning cycle, the replanning period is set to~\SI{2}{\second}.
Depending on the number of vehicles present, the RL planner usually takes no longer than one second to derive a new cooperative maneuver plan.

\begin{figure}
	\centering
	\includegraphics[width=\linewidth]{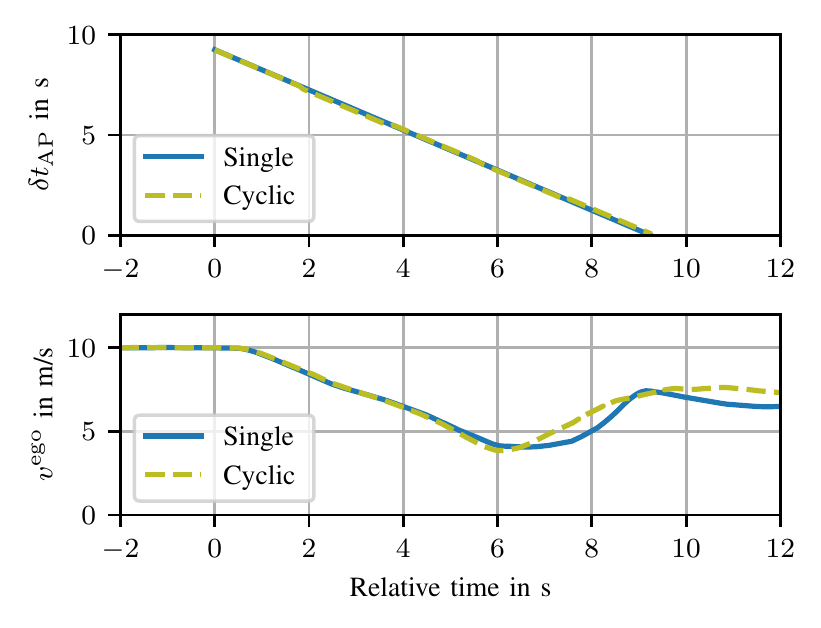}
	\caption{Sample real-world recording of a well manageable maneuver. The deviation between single and cyclic modes is marginal in the anchor timing and insignificant in the velocity profile. The two recordings are depicted relative to the time point in which the RL planner takes over control.}
	\label{fig:simple_example}
\end{figure}

In a first experiment, both vehicles approach the intersection at \SI{10}{\meter\per\second}, which corresponds to the lane speed limit.
By selecting the initial positions $s_0^\mathrm{obj} \approxeq s_0^\mathrm{ego}$ as approximately \SI{-75}{\meter}, a collision would occur if both vehicles remained in constant velocity.
The simulated object vehicle takes priority over the AD vehicle and as such crosses the intersection without braking.
As soon as the RL policy receives an observation of the object vehicle, a decelerating plan is derived for the ego vehicle to cross the intersection behind.
In this case, the behavior of the object vehicle can be anticipated well, which results in a feasible maneuver even when planned in single-shot mode.
Thus, it is fundamentally possible to link an RL policy to a dedicated motion planner through advance planning in the built-in simulator.
As can be observed in Fig.~\ref{fig:simple_example}, the deviation to a cyclic planned maneuver is marginal, which shows the consistency of the RL-based plan.
There are noticeable differences in the speed profile, which can be explained by the internal cost function optimizer of the motion planner.

\begin{figure}
	\centering
	\includegraphics[width=\linewidth]{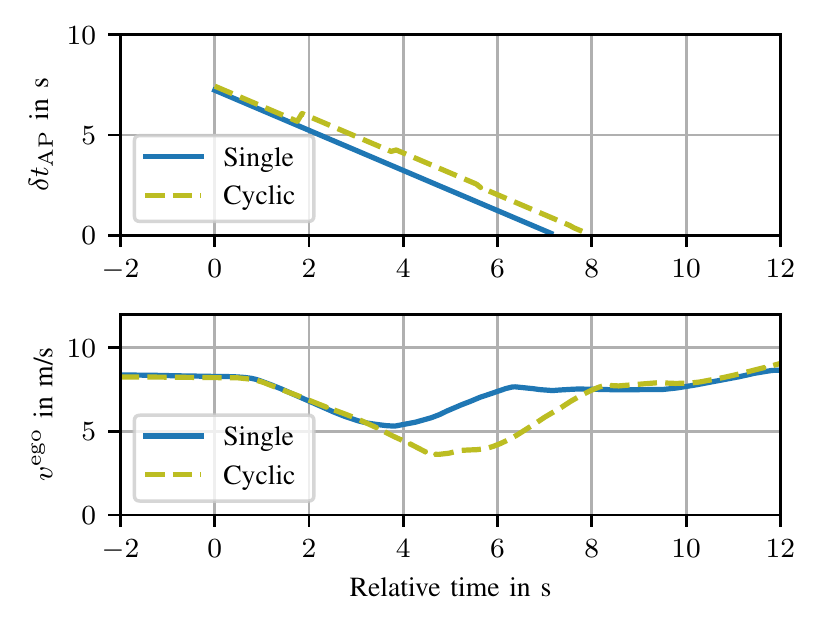}
	\caption{Sample real-world recording of a challenging maneuver. Due to the prioritized crossing vehicle driving slower than anticipated, cyclic replanning increases the anchor time and thus postpones the intersection traversal. The single-shot planning mode, however, cannot perceive this issue.}
	\label{fig:challenge_example}
\end{figure}

In practice, human drivers might reduce the speed of their vehicle at an intersection, even if they have right of way.
Therefore, we conducted a second experiment in which the object vehicle's speed is set to $v^\mathrm{obj} \approxeq \SI{8}{\meter\per\second}$.
During training and in the built-in simulation environment, human-driven vehicles are assumed to accelerate to the lane speed limit whenever possible.
Thus, vehicles going slower without apparent reason might pose a challenge for the RL planner or at least when employing the single planning mode.
Moreover, the take-over point is reduced to $s_0^\mathrm{ego} \approxeq \SI{-55}{\meter}$, and the approach speed is selected as $v_0^\mathrm{ego} \approxeq \SI{8}{\meter\per\second}$, to keep the conflict point with the crossing vehicle.
Indeed, the anchor time and velocity profile in single and cyclic mode exhibit significant deviations, as depicted in Fig.~\ref{fig:challenge_example}.
While the ego vehicle decelerates, cyclic replanning yields two increases in anchor time, which leads to a delay of the intersection traversal of about one second.
The velocity profile analogously shows a further reduction for the cyclic mode, which accommodates the need to delay the vehicle before it arrives at the conflict point.
Nonetheless, the velocity profile remains smooth and, due to the early slowdown, stopping is not required.

\begin{figure}
	\centering
	\includegraphics[width=\linewidth]{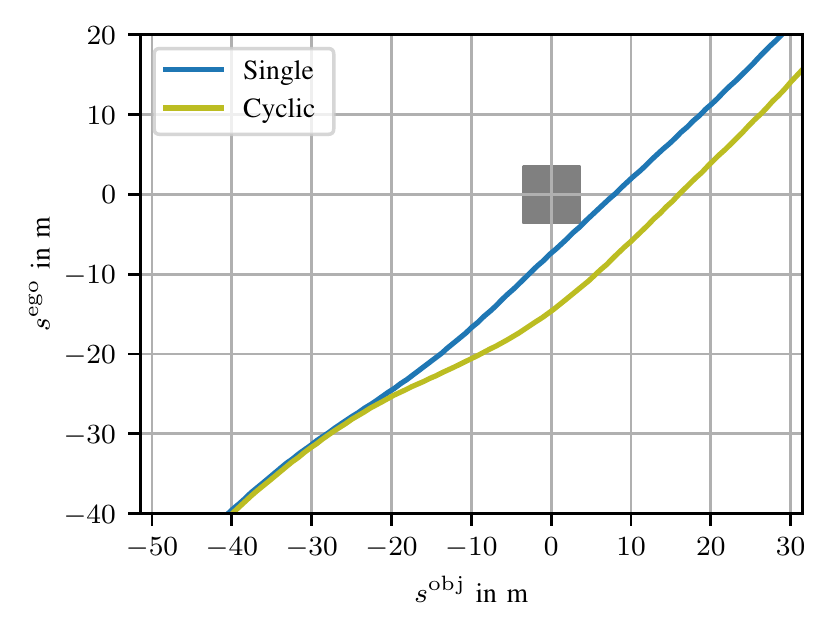}
	\caption{Relative ego vehicle motion over the object vehicle position. The conflict point is located at the origin and the gray square denotes the area in which a collision occurs for assumed vehicle dimensions of $\SI{5}{\meter}\times\SI{2}{\meter}$.}
	\label{fig:relative_s}
\end{figure}

The significance of this mechanism becomes apparent in Fig.~\ref{fig:relative_s} that shows the motion of the ego vehicle relative to the crossing object vehicle's position.
Both coordinates refer to the coordinate frame in Fig.~\ref{fig:maneuver} with its origin on the conflict point.
Thus, the closer a trajectory comes to the origin, the smaller the safety margins between the vehicles.
Assuming a vehicle dimension of $\SI{5}{\meter}\times\SI{2}{\meter}$, the gray square denotes the area in which a collision would occur.
Apparently, the single-shot maneuver is on the edge of colliding and violates all safety constraints.
Cyclic replanning enables a smooth evasive action by slowing down the ego vehicle in time and accelerating once the intersection is clear.
When the object vehicle crosses the conflict point $s^\mathrm{obj} = \SI{0}{\meter}$, the ego vehicle is at $s^\mathrm{ego} \approxeq \SI{-15}{\meter}$, which seems to be a reasonable safety distance.

Although the cyclic replanning algorithm imposes a higher computational load and might lead to minor comfort impairment, it extends the applicability of the RL planner significantly.
In practice, it is most likely inevitable due to numerous effects that are difficult to anticipate, especially under ambiguous drivers' behavior in mixed traffic.

\subsection{Simulative Multi-Agent Planning}
\label{ssec:simulation}

To investigate the suitability of our approach for more dense traffic scenarios, we complement the real-world examples by simulative experiments in cooperative multi-agent planning.
The simulator employs a single-track vehicle model and is virtually exchangeable for the conducted analyses.
This evaluation focuses on cooperative maneuver in fully automated traffic, while each simulated vehicle runs an instance of the same sampling-based motion planner as applied to our testing vehicle in Sec.~\ref{ssec:replanning}.
Analogue to the real-vehicle experiments, each scenario is conducted in the idealistic \emph{single} planning mode and in \emph{cyclic} replanning.

By comparing the traffic scene evolvement between single-shot and cyclic planning, the consistency of the maneuvers planned by the RL policy shall be assessed.
In total, 40 scenarios were generated randomly, each describing an initial configuration of vehicles on the access lanes to the intersection depicted in Fig.~\ref{fig:ap_placement}.
This intersection layout corresponds to the pilot site for automated driving in Ulm-Lehr, Germany~\cite{buchholz2021handling}.
On each lane, one or two vehicles are sampled at a distance of \SI{40}{\meter} to \SI{60}{\meter} to the intersection, leading to a total of three to six vehicles per scenario.
The initial velocities are sampled between \SI{70}{\percent} and \SI{100}{\percent} of the respective lane speed limit.

\begin{figure}
	\centering
	\includegraphics[width=\linewidth]{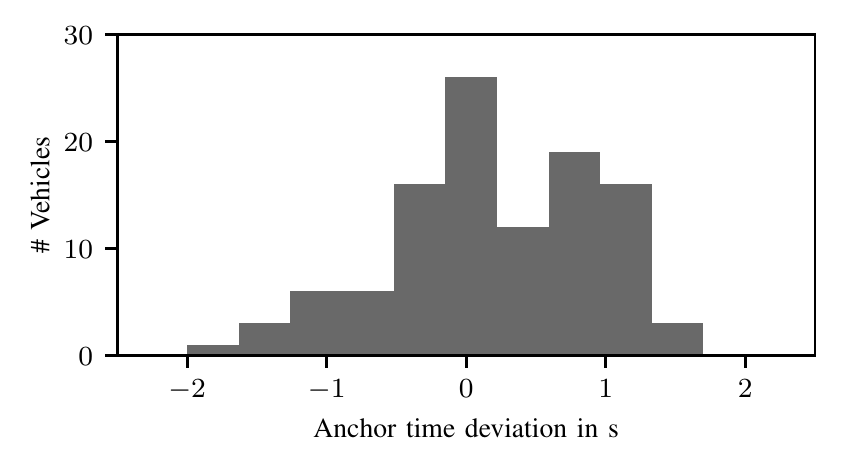}
	\caption{Histogram of the relative anchor time deviation between single-shot planning and cyclic replanning. Scenarios with deviating vehicle crossing order are excluded.}
	\label{fig:anchor_delta}
\end{figure}

Out of those 40 scenarios, 33 exhibit the same vehicle crossing order on the intersection for both planning modes.
In seven scenarios, the crossing order changed during replanning, but not necessarily for all vehicles involved.
We investigate these cases later and first focus on the consistent maneuvers.

The histogram in Fig.~\ref{fig:anchor_delta} depicts the distribution of relative anchor time deviations for the cases of matching vehicle crossing order.
It can be observed that nearly all anchor points remain within \SI{1.5}{\second} of their initially assigned point in time.
These maneuver modifications can be attributed to adaptations of the maneuver to evade collisions with vehicles, whose behavior was not perfectly anticipated, like discussed in Sec.~\ref{ssec:replanning}.
Moreover, minor adaptations of the maneuver might become necessary due to effects of cost function optimization in motion planning and imperfect trajectory tracking.

Reasons for a significant maneuver change during replanning (e.g., change of crossing order) are manifold.
The sampling-based motion planner independently follows a lead vehicle if present.
In this case, a possibly requested anchor point is superseded and instead a follow trajectory is sampled.
This feature prevents the vehicle from a rear-end collision with its lead vehicle in attempt to hit an anchor point, but might have unintended side-effects.
The RL policy might undercut the nominal follow distance, selected by the motion planner, which subsequently leads to a deviation from the high-level plan.
Ultimately, parts of the maneuver might become infeasible if the vehicle in follow mode arrives at the intersection much later than initially anticipated.
A major maneuver revision is then inevitable.
Cyclic replanning enables the RL policy to become aware of such issues and react early on.

\begin{figure}
	\centering
	\includegraphics[width=\linewidth]{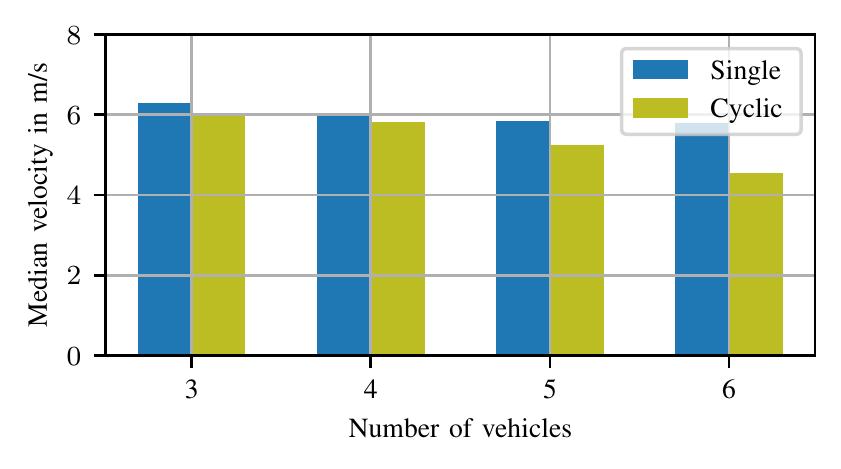}
	\caption{Median driving velocity for planning in single and cyclic mode over a varying number of vehicles in the scenario.}
	\label{fig:velocity}
\end{figure}

The analyses discussed so far suggest that cyclic replanning is more suited and potentially the only viable option for complex or mixed traffic scenarios.
Finally, it shall be assessed whether cyclic replanning comes with a performance degradation compared to the single planning mode.
Figure~\ref{fig:velocity} depicts a bar chart of the median driving velocity over all simulated vehicle trajectories.
The categorization over the number of vehicles in the scene serves as a proxy for the complexity of the traffic scene.
As excepted, the average driving velocity decreases with increasing scene complexity, which is more pronounced for the cyclic mode.
In single planning mode, there is no measure to detect maneuvers becoming infeasible throughout execution, though.
As discussed above, employing cyclic replanning solves these issues, that become more pressing with increasing scene complexity.
For the simple scenarios, comprised of up to four vehicles, the difference between single and cyclic mode is marginal.

\begin{figure}
	\centering
	\includegraphics[width=\linewidth]{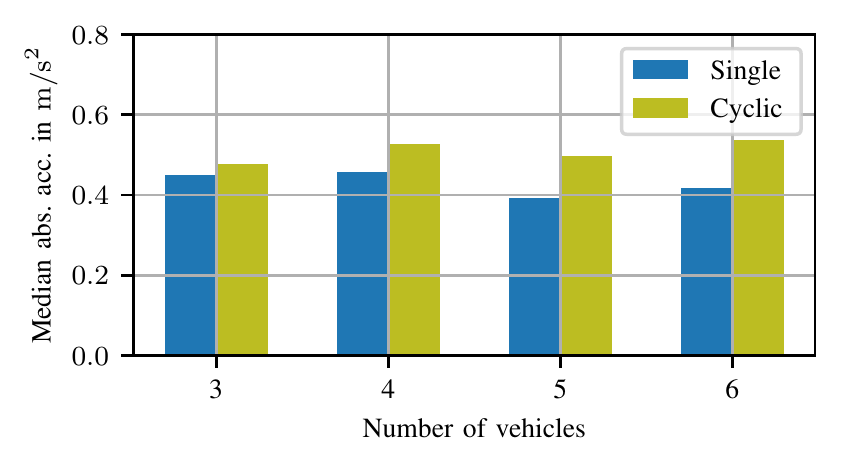}
	\caption{Median absolute acceleration for planning in single and cyclic mode over a varying number of vehicles in the scenario.}
	\label{fig:acceleration}
\end{figure}

Running the RL policy in cyclic replanning inherently leads to a more agile behavior, especially in case of maneuver changes.
Thus, the average absolute acceleration increases moderately compared to single-shot planning, as it can be observed in Fig.~\ref{fig:acceleration}.
Again, the difference is more apparent for the complex maneuvers.
Although single-shot planning seems desirable in terms of velocity and acceleration statistics, its application in practice is unrealistic because of serious safety concerns.
Out of the 40 simulated scenarios, six single-shot maneuvers end in a collision of two or more involved vehicles.
Cyclic replanning remains collision-free.
Here, a collision is defined as the case that the bounding boxes of vehicles intersect for at least one time step.
We conclude that cyclic replanning is required to effectively plan feasible non-trivial maneuvers consisting of, e.g., multiple vehicles per lane or in mixed traffic.

\section{Conclusion}
\label{sec:conclusion}

This work proposed a novel approach to employ an RL policy dedicatedly for behavior planning in automated driving.
In contrast to prior art, we do not perform end-to-end planning, but supplement a well-tuned sampling-based motion planner.
Thereby, a built-in simulation environment is used which rolls out the traffic scene by querying the RL policy.
The simulated trajectories are subsequently turned into viable MP objectives for detailed trajectory planning.
Due to lack of benchmark and baseline approaches, we evaluated our approach by comparing an idealistic single-shot plan with cyclic replanning through the RL objective.
By conducting experiments on a real testing vehicle, the algorithm's potential to shrink the RL simulation to real world gap through cyclic replanning was demonstrated.
Additionally, simulative evaluation showed the fundamental applicability of this method to more complex scenarios in cooperative multi-agent planning.

\balance
\bibliographystyle{IEEEtran}
\bibliography{references}

\end{document}